\documentclass{article}
\PassOptionsToPackage{numbers, compress}{natbib}

\usepackage[preprint]{neurips_2020}

\usepackage{wrapfig,lipsum,booktabs}

\usepackage[utf8]{inputenc} % allow utf-8 input
\usepackage[T1]{fontenc}    % use 8-bit T1 fonts
\usepackage{lmodern}
\usepackage{hyperref}       % hyperlinks
\usepackage{url}            % simple URL typesetting
\usepackage{booktabs}       % professional-quality tables
\usepackage{amsfonts}       % blackboard math symbols
\usepackage{nicefrac}       % compact symbols for 1/2, etc.
\usepackage{microtype}      % microtypography
\usepackage{subfig}
\usepackage{amsmath} 
\usepackage{amssymb} 
\usepackage{array}
\usepackage{float}
\usepackage{enumitem}
\usepackage{graphicx}
\usepackage{natbib} 
\usepackage{booktabs} 
\usepackage{easytable}
\usepackage{syntonly}
\usepackage{layout}
\usepackage{pstricks}
\usepackage{tikz}
\usepackage{natbib}
\usepackage{color,soul}
\usepackage[ruled,vlined]{algorithm2e}

\title{Accelerating training in artificial neural networks with dynamic mode decomposition}

\author{%
  Mauricio E. Tano  \\
  Nuclear Engineering Department\\
  Texas A\&M University\\
  \texttt{mtano@tamu.edu} \\
  \and
  Gavin D. Portwood \\
  Los Alamos National Laboratory\\
  Los Alamos, NM 87545\\
  \texttt{portwood@lanl.gov} \\
  \and
  Jean C. Ragusa \\
  Nuclear Engineering Department\\
  Texas A\&M University\\
  \texttt{jean.ragusa@tamu.edu} \\
}

\begin{document}

\maketitle

%\vspace{-5em}
\begin{abstract}
Training of deep neural networks (DNNs) frequently involve optimizing several millions or even billions of parameters. Even with modern computing architectures, the computational expense of DNN training can inhibit, for instance, network architecture design optimization, hyper-parameter studies, and integration into scientific research cycles.  The key factor limiting performance is that both the feed-forward evaluation and the back-propagation rule are needed for each weight during optimization in the update rule. In this work, we propose a method to decouple the evaluation of the update rule at each weight, based on the hypothesis that the global evolution of weights at each layer can be modeled by a reproducing Hilbert space. At first, Proper Orthogonal Decomposition (POD) is used to identify a current estimate of the principal directions of evolution of weights per layer during training based in the evolution observed with a few backpropagation steps. Then, Dynamic Mode Decomposition (DMD) is used to learn the dynamics of evolution of the weights in each layer according to these principal directions. The DMD model is used to evaluate an approximate converged state when training the ANN. Afterwards, some number of backpropagation steps are performed, starting from the DMD estimates, leading to an update to the principal directions and DMD model. This iterative process is repeated until convergence. By fine-tuning the number of backpropagation steps used for each DMD model estimation, a significant reduction in the number of operations required to train neural network can be achieved. In this paper, the DMD acceleration method will be explained in detail, along with the theoretical justification for the acceleration provided by DMD. This method is illustrated using a regression problem of key interest for the scientific machine learning community: the prediction of a pollutant concentration field in a diffusion, advection, reaction problem.
\end{abstract}

%%%%%%%%%%%%%%%%%%%%%%%%%%%%%%%%%%%%%%%%%%%%%%%%%%%%%%%%%%%%%%%%%%%%%
\section{Introduction}
%%%%%%%%%%%%%%%%%%%%%%%%%%%%%%%%%%%%%%%%%%%%%%%%%%%%%%%%%%%%%%%%%%%%%
In the last five years, machine learning methods have paved it way in science and engineering.
For example, machine learning methods have been proposed to develop turbulence models \cite{tracey2015,portwood19}, to control nonlinear dynamical systems~\cite{duriez2017}, to optimize radiation doses on oncology patients~\cite{saga2015}, and to solve large-scale radiation transport problems~\cite{tano2019}, between many others.
In most cases, one is interested in solving large-scale regression problems when applying machine learning to physical systems.
These problems often involve large Deep Neural Networks (DNNs).
However, the long time required to train these neural networks, even in multi-GPU/TPU architectures, inhibits the possibility of optimizing the architecture of neural networks and its hyperparameters, which are key for integrating these networks into scientific research cycles.

%%%%%%%%%%%%%%%%%%%%%%%%%%%%%%%%%%%%%%%%%%%%%%%%%%%%%%%%%%%%%%%%%%%%%
\section{Motivation}
%%%%%%%%%%%%%%%%%%%%%%%%%%%%%%%%%%%%%%%%%%%%%%%%%%%%%%%%%%%%%%%%%%%%%
Different methods have been recently proposed to accelerate training.
On one hand, {\it instance shrinking methods} accelerate training by filtering out instances that show relatively small marginal improvements, see~\cite{zhang2019}, for example.
However, in nonlinear regression problems, filtering out instances has a big impact on the performance of the model.
Hence, these approaches have not been further explored.
Another family of approaches to accelerate training in DNNs is {\it weight extrapolation}.
In this one, one tries to infer the path of evolution of weights and use it to speed up training.
For this purpose, \cite{kamarthi1999} proposed a line-search procedure to estimate the converged values of each weight during training.
However, when applied to large DNNs, errors in the line-search procedure break the coherent dynamics of the evolution of weights at each layer~\cite{hoskins2019}.
Thus, line-search procedures can be ineffective at accelerating training.
Another approach consists of learning the dynamics of the evolution of weights in a large DNN with a surrogate neural network~\cite{sinha2017}.
Then, this surrogate is used to accelerate training extrapolating weights.
However, even in asynchronous training scenarios, the surrogate model only learned an appropriate performance once the large neural network has been trained.

\begin{figure}
    \centering
    \includegraphics[scale=0.45]{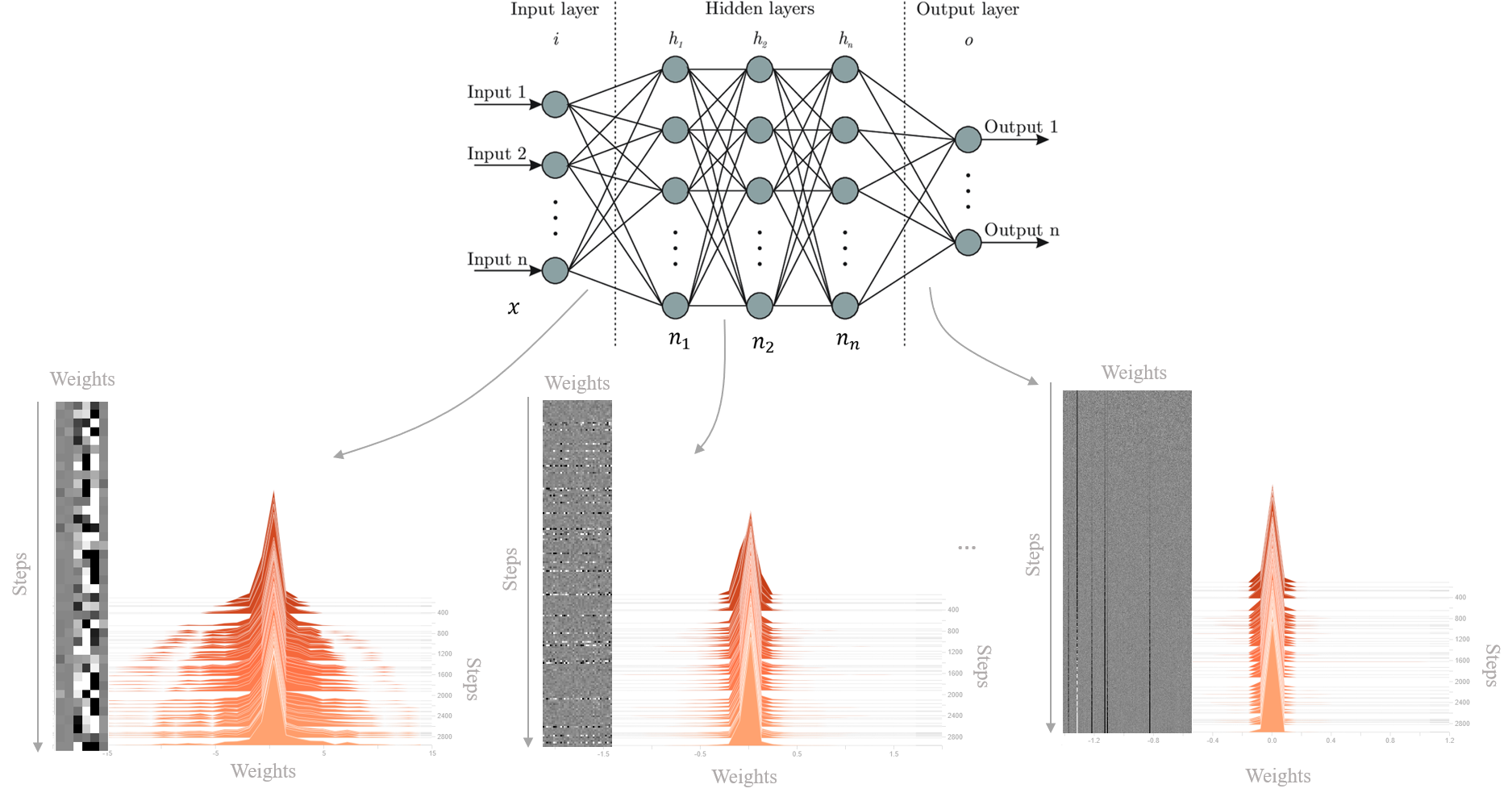}
    \caption{Feed-forward deep neural network architecture considered a visualization of the evolution weights as a function of backpropagation steps.}
    \label{fig:key_idea}
\end{figure}

Hence, more effective acceleration methods are yet necessary to tackle regression problems.
In this work, we propose to use Dynamic Mode Decomposition (DMD, see \cite{schmid2010} for a comprehensive review) to learn the dynamics in the evolution of weights per layer.
The key motivation is illustrated by analyzing Figure~\ref{fig:key_idea}.
The Figure presents a DNN with three hidden-layers that has been trained to predict the concentration of a pollutant in the atmosphere as a function of six input parameters.
These input parameters are the uncertain parameter that describe the physical problem.
Weights have been initialized using the Xavier initializer~\cite{glorot2010} and the DNN is trained with the Adam algorithm~\cite{adam}.
Three interesting things are observed.
First, the distributions of weights follow a monotonic evolution as optimization steps are performed.
This indicates an underlying dynamic in the evolution of weights.
Second, when tracking the evolution of weights per layer, pikes and deeps are observed coherently for all weights.
This indicates that the dynamics in the evolution of weights must be analyzed in batches per layer.
A similar conclusion to this last one motivated~\cite{hoskins2019}.
Finally, a noisy evolution is observed on the weights.
This noise extends beyond the stochastic noise introduced in the Adam optimizer.
Hence, some sort of filter should be necessary to accelerate training.
By selecting only a few principal modes a filter can be embedded in DMD.

The rest of this paper is organized as follows. 
In the next section, a low computational-cost version of DMD is described, as well as the algorithm by which DMD is embedded to accelerate the training process.
Then, the extent to which DMD can reproduce the weight evolution dynamics in a DNN is analyzed along with the reduction in computational complexity provided by this algorithm.
Finally, the application of the method to a DNN solving the problem of a transporting reacting pollutant in the atmosphere is analyzed.

%%%%%%%%%%%%%%%%%%%%%%%%%%%%%%%%%%%%%%%%%%%%%%%%%%%%%%%%%%%%%%%%%%%%%
\section{Enhanced training with Dynamic Mode Decomposition}
%%%%%%%%%%%%%%%%%%%%%%%%%%%%%%%%%%%%%%%%%%%%%%%%%%%%%%%%%%%%%%%%%%%%%
A low computational-cost version of DMD is introduced first and the enhanced training algorithm afterwards.

Let $\boldsymbol{W}^{\ell,m} \in \mathbb{R}^{(\ell -1) \ell  \times m}$ be the matrix obtained by flattening the weights for layer $\ell$ and taking $m$ steps of the optimizer, i.e., a truncated version of the matrices presented in Figure~\ref{fig:key_idea}.
This matrix is generally called {\it snapshots matrix} within the DMD approach.
We note that the dimension the rows in these matrices, referred to as $n = (\ell -1) \ell$ for brevity, is much smaller that the dimensions of the columns, this is $n \gg m$.
The first step of DMD consists of taking the Singular Value Decomposition (SVD) of the weight matrices per layer as follows:
\begin{equation*}
    \boldsymbol{W}^{\ell,m} = \boldsymbol{U}^{\ell,m} \boldsymbol{\Sigma}^{\ell,m} {\boldsymbol{V}^{\ell,m}}^T \,,
\end{equation*}
where $\boldsymbol{U}^{\ell,m}$ is a matrix containing the batch modes for weights, $\boldsymbol{\Sigma}^{\ell,m}$ is a diagonal matrix containing the amplification factors of these modes, and ${\boldsymbol{V}^{\ell,m}}^T$ is a matrix that contains the directions of evolution of the modes during the training steps.

The computational complexity of classical SVD is $\mathcal{O}(n^2m)$, which is large considering the dimension of $n$.
However, a cheaper version can be obtained by performing SVD on the columns of $\boldsymbol{W}^{\ell,m}$ instead of its rows.
This is, we build the product ${\boldsymbol{W}^{\ell,m}}^T \boldsymbol{W}^{\ell,m}$ which is of order $\mathcal{O}(nm^2)$.
By definition of SVD,
\begin{equation*}
    {\boldsymbol{W}^{\ell,m}}^T \boldsymbol{W}^{\ell,m} = \boldsymbol{V}^{\ell,m} {\boldsymbol{\Sigma}^{\ell,m}}^2 {\boldsymbol{V}^{\ell,m}}^T \text{ ,}
\end{equation*}
which is an eigenvalue problem that can be solved with a computational complexity of $\mathcal{O}(m^3 \ll nm^2)$.
Then, the matrix $\boldsymbol{U}^{\ell,m}$ can be reconstructed by taking the products 
\begin{equation*}
\boldsymbol{U}^{\ell,m} =  \boldsymbol{W}^{\ell,m}\boldsymbol{V}^{\ell,m} {\boldsymbol{\Sigma}^{\ell,m}}^{-1} \text{ ,}
\end{equation*}
which can be performed with a computational complexity of $\mathcal{O}(nm^2)$.
Thus, all computations can be performed with a computational complexity of $\mathcal{O}(nm^2)$.

Then, the key idea of DMD is to build a reduced version of the Koopman operator, which is a matrix linearly relating the states of the weights.
In this sense, the Koopman operator $\boldsymbol{A}^{\ell,m}$ allows relating the flattened weights at step $m+1$, $\boldsymbol{w}^{\ell,m+1}$, with the ones at step $m$, $\boldsymbol{w}^{\ell,m}$, as  $\boldsymbol{w}^{\ell,m+1} = \boldsymbol{A}^{\ell,m} \boldsymbol{w}^{\ell,m}$.
However, note that in this definition, the Koopman operator will be subjected to the random noise during training obtained for the weights.
To solve  this problem, we define generalized matrices.
A lagged matrix ${\boldsymbol{W}^{\ell,m}}^-$, which take the first $m-1$ columns of $\boldsymbol{W}^{\ell,m}$, and a forwarded matrix ${\boldsymbol{W}^{\ell,m}}^+$, which take the last $m-1$ columns of $\boldsymbol{W}^{\ell,m}$.
Then, we build a generalized version of the Koopman operator as ${\boldsymbol{W}^{\ell,m}}^+ = \boldsymbol{A}^{\ell,m} {\boldsymbol{W}^{\ell,m}}^-$, which should be less subjected to noise in virtue the larger stencil of columns used.
Following, we introduce the SVD decomposition of the lagged matrix 
\begin{equation}
{\boldsymbol{W}^{\ell,m}}^- = \boldsymbol{U}_r^{\ell,m} \boldsymbol{\Sigma}_r^{\ell,m} {\boldsymbol{V}_r^{\ell,m}}^T 
\label{eq:lmatm}
\end{equation}
where the subscript $r$ indicates that only the first $r<m$ modes have been selected for the weights (effectively filtering noise out of the system), yielding 
\begin{equation}
{\boldsymbol{W}^{\ell,m}}^+ = \boldsymbol{A}^{\ell,m} \boldsymbol{U}_r^{\ell,m} \boldsymbol{\Sigma}_r^{\ell,m} {\boldsymbol{V}_r^{\ell,m}}^T \text{ .} 
\label{eq:lmatp}
\end{equation}
A reduced version of the Koopman operator can be built by its projection into the weights modes, i.e., $\boldsymbol{A}_r^{\ell,m} = {\boldsymbol{U}_r^{\ell,m}}^T \boldsymbol{A}^{\ell,m} \boldsymbol{U}_r^{\ell,m}$. 
Seeking for the reduced version of the Koopman in its previous definition yields, 
\begin{equation}
\boldsymbol{A}_r^{\ell,m} = {\boldsymbol{U}_r^{\ell,m}}^T {\boldsymbol{W}^{\ell,m}}^+ \boldsymbol{V}_r^{\ell,m} {\boldsymbol{\Sigma}_r^{\ell,m}}^{-1}\text{ .}
\label{eq:rkoop_gen}
\end{equation}
The computational complexity of building the reduced Koopman operator is $\mathcal{O}(nm^2+2m^3) \sim \mathcal{O}(nm^2)$.
Then, the dynamics embedded in the reduced Koopman operator can be extracted by solving the eigenvalue problem \begin{equation}
\boldsymbol{A}_r^{\ell,m} \boldsymbol{Y}_r^{\ell,m} = \boldsymbol{\Lambda}_r^{\ell,m} \boldsymbol{Y}_r^{\ell,m} \text{ ,}
\label{eq:eig_rkoop}
\end{equation} which has a computational complexity of $\mathcal{O}(r^3)$.
Note that $\boldsymbol{\Lambda}_r^{\ell,m}$ dictates the dynamic of evolution of weights step-by-step, while $\boldsymbol{Y}^{\ell,m}$ specifies how the principal modes for the weights are combining in this dynamics defining matrix of modes $\boldsymbol{\Phi}_r^{\ell,m} = {\boldsymbol{U}_r^{\ell,m}}^+ \boldsymbol{Y}^{\ell,m}$.
Another $\mathcal{O}(nr^2)$ are needed to compute ${\boldsymbol{U}_r^{\ell,m}}^+$ and the product building $\boldsymbol{\Phi}_r^{\ell,m}$
Hence, the dynamics of the evolution of the weights at a layer $\ell$ for a step $s>m$ are given by:
\begin{equation}
    \boldsymbol{w}^{\ell,s} = \boldsymbol{\Phi}_r^{\ell,m} \left[ \boldsymbol{\Lambda}_r^{\ell,m} \right]^{s-m} \boldsymbol{b}_r^{\ell,m} \,,
    \label{eq:update_dmd}
\end{equation}
where $\boldsymbol{b}_r^{\ell,m}$ are the initial coefficients upon which DMD is computed, which are computed as $\boldsymbol{b}_r^{\ell,m} = {\boldsymbol{\Phi}_r^{\ell,m}}^T \boldsymbol{w}^{\ell,s}$.

Note that the update rule is performed at each layer $\ell$ independently of other layers and, thus, can be easily parallelized.
Summarizing, a total of $\sim n \times (3m^2+r^2)$ operations are necessary to compute DMD.
In backpropagation, the computational complexity is of order $\mathcal{O}(nt)$, where $t$ is the number of samples in the training set.
Therefore, DMD will provide acceleration as long as $t < 3m^2+r^2$.
This condition is generally achieved given the large number of samples $t$ used in regression training sets.
Nonetheless, we note that the computational complexity grows with $m^2$ and $r^2$.
So, the algorithm will be less efficient if a large number of steps $m$ are included in the matrix of weights or as more modes $r$ are retained when performing the DMD evolution.

The complete algorithm on the usage of DMD to accelerate backpropagation is presented in Algorithm~\ref{algo:DMDpBP}.
Initially, the number of steps to be included in the weight matrices for all layers ($m$), a filtering tolerance on the singular values (which determine $r$), and the number of steps on which DMD will extrapolate ($s$) must be provided by the user, along with classical backpropagation parameters (e.g., learning rate, momentum, etc.).
Then, classical backpropagation steps are performed, while building the snapshot matrices for weights.
When the number of backpropagation steps equals the limit $m$, the DMD loop is triggered.
In this one, the DMD process previously described is carried out and the new weights computed with DMD are assigned to the DNN model per layer.
Next, the number of backprogation iterations to begin the DMD loop is set to zero and the process is repeated.
Note that the whole for loop in this algorithm can be easily parallelized by computing DMD modes and updating weights concurrently across all layers.

\begin{algorithm}%[H]
\SetAlgoLined
\KwIn{$m$, DMD filter tolerance, $s$, Backpropagation Parameters, Total Epochs}
\KwResult{Trained weights $\boldsymbol{w}^{\ell}, \forall \ell \in \mathcal{H}_\ell$}
 $bp_{iter} = 0$,\; 
 \While{epoch $\leq$ Total Epochs}{
  Do backpropagation step\;
  Extract weights $\boldsymbol{w}^{\ell, bp_{iter}} \,, \forall \ell \in \mathcal{H}_\ell$\;
  Store weights: $\boldsymbol{W}^{\ell} \leftarrow [\boldsymbol{W}^{\ell} \, \boldsymbol{w}^{\ell, bp_{iter}}] \,, \forall \ell \in \mathcal{H}_\ell$\;
  $bp_{iter} += 1$ \;
  \If{$bp_{iter} == m$}{
   \For{$\ell \in \mathcal{H}_{\ell}$}{
       Build training matrices: ${\boldsymbol{W}^{\ell,m}}^-$ and ${\boldsymbol{W}^{\ell,m}}^+$ with \eqref{eq:lmatm} and \eqref{eq:lmatp}\;
       Compute low-cost SVD decomposition: ${\boldsymbol{W}^{\ell,m}}^- = \boldsymbol{U}_r^{\ell,m} \boldsymbol{\Sigma}_r^{\ell,m} {\boldsymbol{V}_r^{\ell,m}}^T$\;
       Select $r$ modes such that $\boldsymbol{\Sigma}_r^{\ell,m}[r,r] / \boldsymbol{\Sigma}_r^{\ell,m}[0,0] > \text{DMD filter tolerance}$\;
       Build reduced Koopman operator with \eqref{eq:rkoop_gen}\;%: $\boldsymbol{A}_r^{\ell,m} = {\boldsymbol{U}_r^{\ell,m}}^T {\boldsymbol{W}^{\ell,m}}^+ \boldsymbol{V}_r^{\ell,m} {\boldsymbol{\Sigma}_r^{\ell,m}}^{-1}$\;
       Perform eigendecomposition of the reduced Koopman operator with \eqref{eq:eig_rkoop} \;%: $\boldsymbol{A}_r^{\ell,m} \boldsymbol{Y}_r^{\ell,m} = \boldsymbol{\Lambda}_r^{\ell,m} \boldsymbol{Y}_r^{\ell,m}$\;
       Compute the matrix of weights modes: $\boldsymbol{\Phi}_r^{\ell,m} = {\boldsymbol{U}_r^{\ell,m}}^+ \boldsymbol{Y}^{\ell,m}$
       Compute initial DMD condition: $\boldsymbol{b}_r^{\ell,m} = {\boldsymbol{\Phi}_r^{\ell,m}}^T \boldsymbol{w}^{\ell,m}$\;
       Evolve weights with DMD using \eqref{eq:update_dmd} \;%: $\boldsymbol{w}^{\ell,s} = \boldsymbol{\Phi}_r^{\ell,m} \left[ \boldsymbol{\Lambda}_r^{\ell,m} \right]^{s-m} \boldsymbol{b}_r^{\ell,m}$\;
       Assign updated weights to layer $\ell$ in the neural network\;
   }
   $bp_{iter} = 0$\;
   }
 }
 \caption{Acceleration of backpropagation with Dynamic Mode Decomposition}
 \label{algo:DMDpBP}
\end{algorithm}

As previously mentioned, we decided to study the performance of this method for a problem of factual interest for scientific machine learning, which is the dispersion of a pollutant in the atmosphere.
Key details on this problem will be provided in the next section, along with a sensitivity study to the $m$ and $s$ parameters and an analysis of the training of the DNN with and without DMD acceleration.

%%%%%%%%%%%%%%%%%%%%%%%%%%%%%%%%%%%%%%%%%%%%%%%%%%%%%%%%%%%%%%%%%%%%%
\section{Regression problem: dispersion of a reactive pollutant in the atmosphere}
%%%%%%%%%%%%%%%%%%%%%%%%%%%%%%%%%%%%%%%%%%%%%%%%%%%%%%%%%%%%%%%%%%%%%
We now demonstrate the DMD-accelerated training by considering the dispersion of a reactive pollutant in the atmosphere.
A complete mathematical description of the problem is provided in Appendix 1 but some key details are given in this section.
The problem is described as follows.
Two reactants of concentration $c_1$ and $c_2$ are dispersed in the atmosphere from an approximately constant source.
These two react to produce the pollutant $c_3$ with an unknown first-order reaction constant $K_{12}$.
Also, the pollutant disappears with a first-order reaction constant $K_3$.
Both reactants and the pollutant diffuse in the atmosphere with the same diffusion coefficient $D$.
In addition, all three are advected by convective currents in the atmosphere.
The convective current velocity field is determined by three parameters: the wind speed ($U_0$), the slip velocity next to the ground ($u_h$), and the buoyant vertical velocity next to the ground ($u_v$).

\begin{figure}
    \centering
    \includegraphics[scale=0.55]{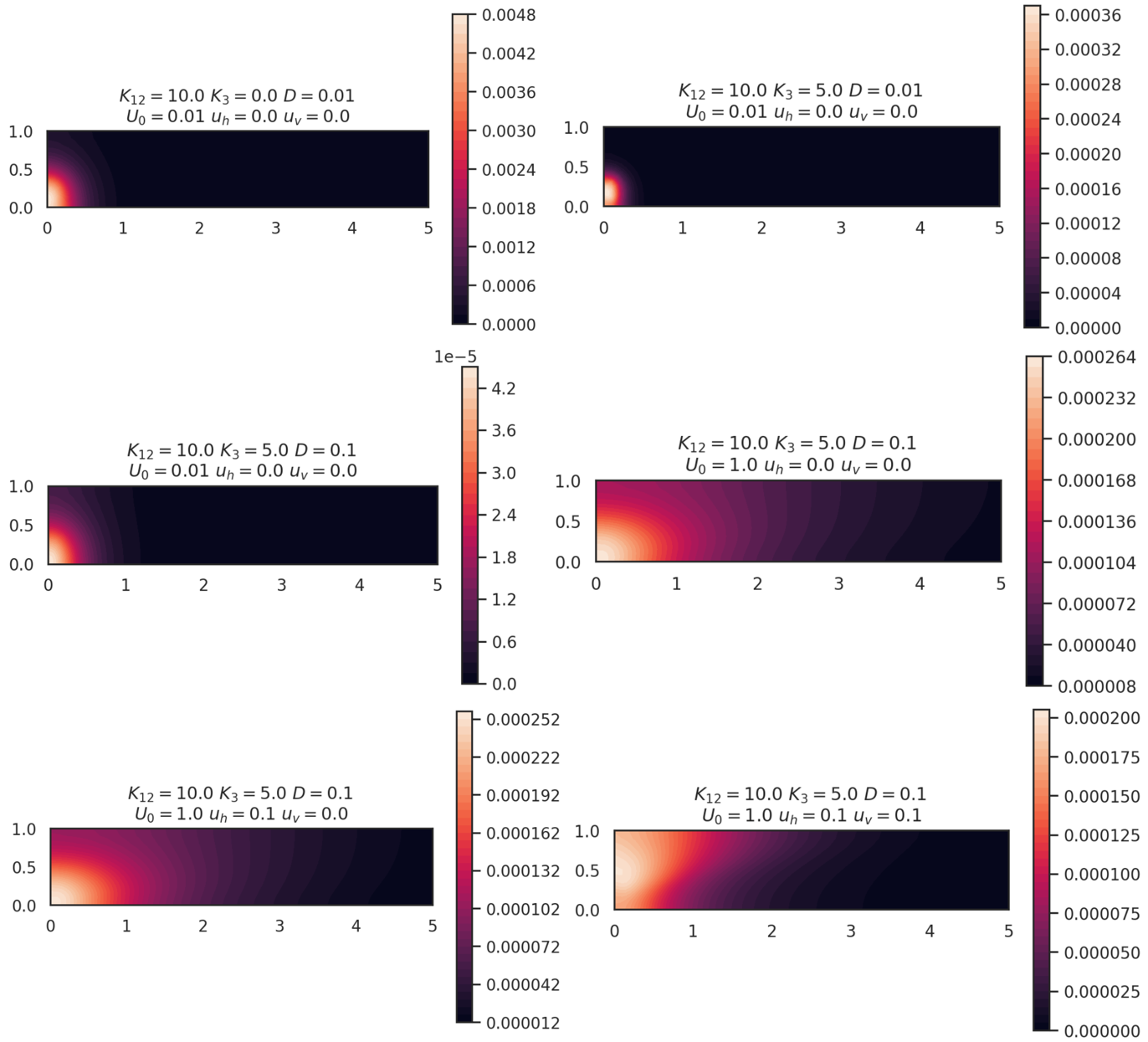}
    \caption{Steady state solutions for the concentration field of the dispersed pollutant changing the uncertain parameters one at the time.}
    \label{fig:ADR}
\end{figure}

In order to understand the effect of the uncertain parameters in the problem, Figure~\ref{fig:ADR} presents steady-state solutions for the concentration field of the pollutant varying one parameter at  the time.
We will analyze each simulation from left to right and from top to bottom.
First, a large production coefficient $K_{12}$ will focus the concentration of pollutant next to the emission source, placed in the bottom left corner of the domain.
The, increasing the decay rate $K_3$ for the pollutant, will attenuate its concentration throughout the domain.
Additionally, increasing the diffusion coefficient $D$, will smooth out the distribution of pollutant next to the emission source.
Following, adding an advection velocity $U_0$, will transport the pollutant from left to right in the domain.
Moreover, increasing the horizontal velocity next to the ground $u_h$ will result in a further advection downstream of the pollutant.
Finally, a positive buoyant velocity $u_v$ next to the ground in the bottom of the domain will result in a transport of the pollutant away from the bottom of the domain.

In the regression problem the concentration of the pollutant at $2670$ points in space (placed preferentially next to the source and next to the bottom plate) should be determined as a function of the six uncertain parameters ($K_{12}, K_3, D, U_0, u_h, u_v$).
For this purpose a deep neural network with three hidden-layers with $40$, $200$, and $1000$ is built, using soft-sign as activation in the hidden layers, which have been trained which GPU acceleration using the Google Colabs infrastructure as in May 2020.
The training data is developed by numerical simulations of the transport problem.
$10^3$ samples of the six uncertain parameters are generated with the Latin Hypercube Sampling method~\cite{stein1987} and the steady state concentration of pollutant is obtained by running a numerical simulation for each parameter set.
$80 \%$ of the data is used for training, while $20 \%$ is added to the testing set for validation purposes.
The sampling ranges are $K_{12} \in [1.0,20.0], K_3 \in [0.0,10.0], D \in [0.01,0.5], U_0 \in [0.01,2.0], u_h \in [-0.2,0.2], u_v \in [-0.2,0.2]$.
Due to this large sampling range we expect large variations in the concentration fields obtained for the polutants.
Both input and output are scaled and normalized to convenient ranges of the activation function before training.
The goal of building this DNN is to have a tool that can rapidly evaluate the concentration of the pollutant and apply a rapid response in case of an emergency.
However, this network has $\sim 2.9 \times 10^6$ trainable parameters.
Hence, its training and optimization can be cumbersome.
For this purpose, the DMD outlined in this manuscript is applied to accelerate training.

To understand the interaction between DMD and backpropagation during training a sensitivity study on the $m$ and $s$ parameters is first developed.
For this purpose Algorithm~\ref{algo:DMDpBP} is run with parameters $m \in [2,20]$ and $s \in [5,100]$.
We focus on the average relative error provided by DMD, i.e., the mean squared error after applying the DMD process divided by the error before applying this one.
Additionally, since several DMD processes will be applied when ruining Algorithm~\ref{algo:DMDpBP}, we define as metric the mean average relative error, which is the unweighted mean of the relative error provided in the DMD processes throughout training.
A total of $3000$ epochs have been performed during training, using the mean square error as loss function and Adam as optimizer.
Hence, the mean relative improvement is computed from $1500$ samples for the case with $m = 2$ to $150$ samples for the case with $m=20$.
A DMD filtering factor of $10^{-10}$ is used, which introduces only a mild filter in the DMD evolution.

\begin{figure}
    \centering
    \includegraphics[scale=0.45]{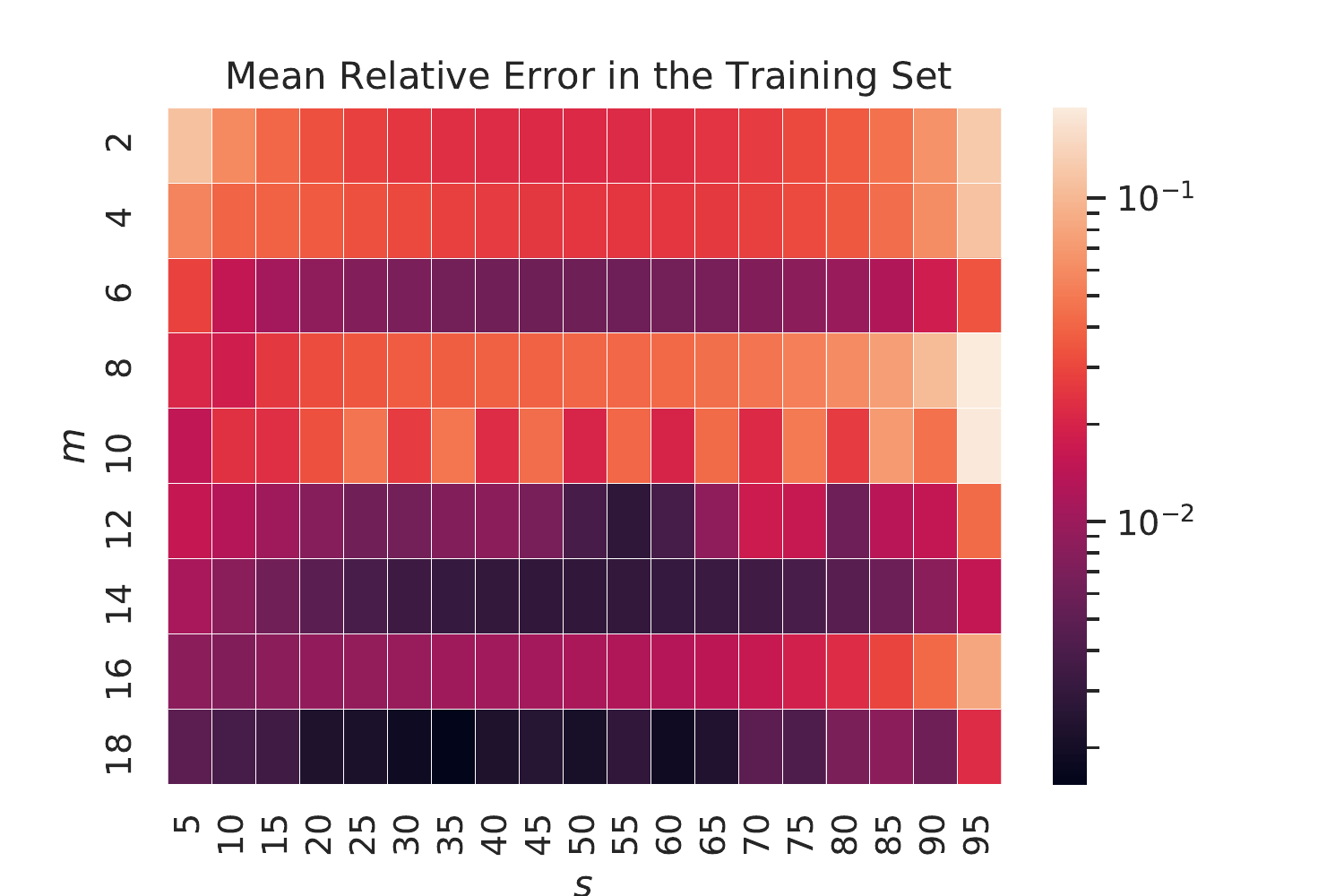}
    \includegraphics[scale=0.45]{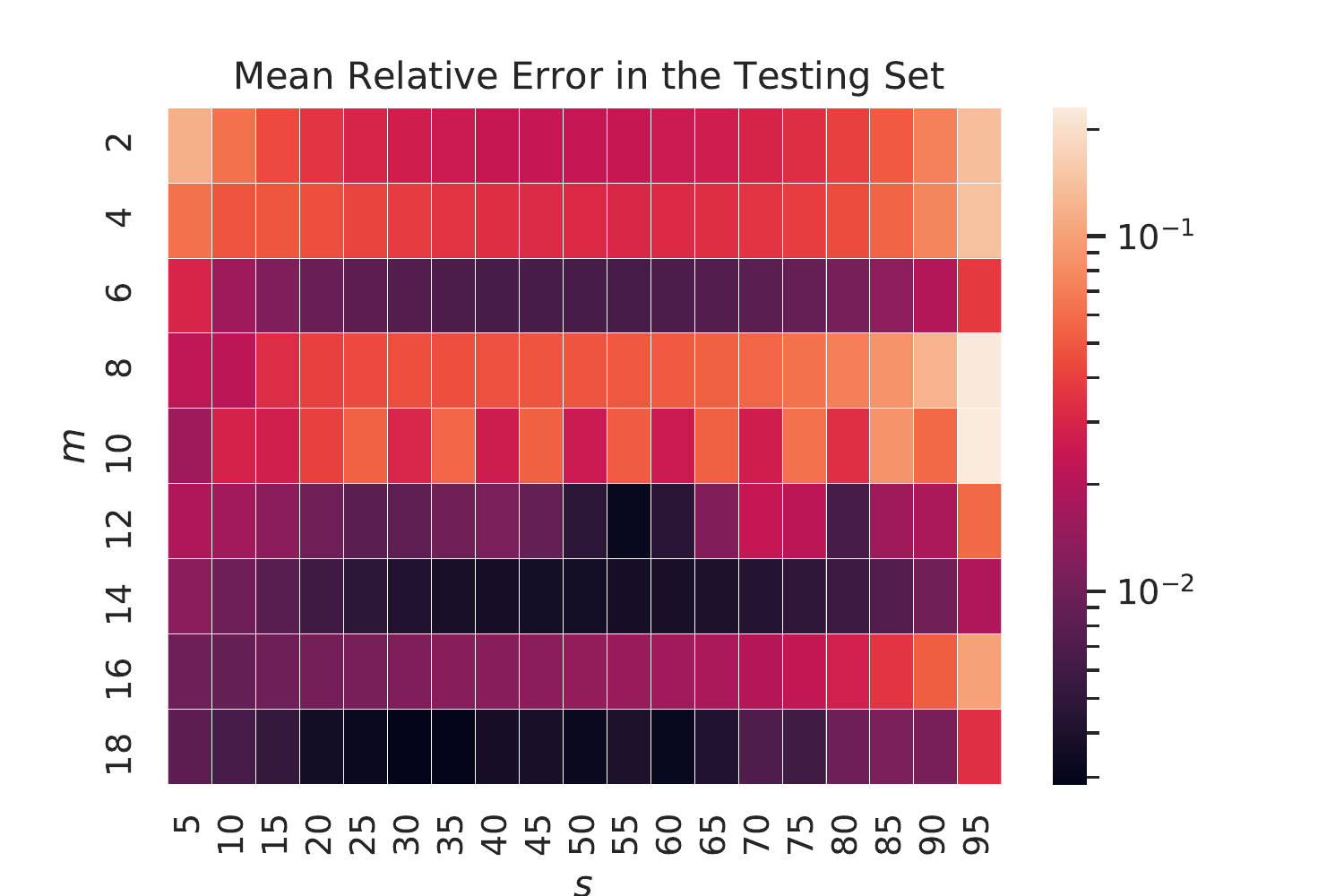}
    \caption{Sensitivity study for the mean relative improvement in each DMD acceleration for the number of snapshots in the weight matrix $m$ and the number of DMD steps in acceleration $s$ for the training set (left) and the testing set (right).}
    \label{fig:TTsensitivity}
\end{figure}

The results obtained from this sensitivity study are presented in Figure~\ref{fig:TTsensitivity}.
Two important factors should be extracted from this figure.
First, the mean relative error decreases as $m$ is larger.
This is expected because the DMD modes are computed with more dynamic information and subject to less noise as $m$ increases.
Furthermore, when taking more modes, the randomness introduced in the stochastic optimization algorithm have a less impact over the DMD modes.
Second, we see that the relative error decreases as more steps $s$ are taken.
This is logical, since DMD is simply advancing towards the predicted converged state of the weights according to the prediction made after the $m$ steps.
However, as more steps $s$ are taken, the relative error start to increase again independently of the number of steps $m$ added in the weight matrices.
This is surprising at first.
However, this is because the DMD process is following a trajectory for the weights determined by the $m$ steps.
Hence, for the first few steps in $s$, it is logical that the loss function decreases since the guess for the gradient at $m$ is still quite accurate.
However, as more steps are taken, the guess for the gradient at $s$ coming from $m$ is evidently less accurate.
Thus, it is logical that the error increases.

Further analysis of Figure~\ref{fig:TTsensitivity} show that approximately the same behaviour that has been observed for the training set is observed for the testing set.
This is important, since it shows that the DMD iterations are not leading the system to over-fitting.
Finally, we can use this sensitivity study to determine the number of steps $m$ to include in the snapshot weight matrices and the number of steps $s$ to take in the DMD iterations.
From Figure~\ref{fig:TTsensitivity} it can be observed that the smallest relative error are produced for $m = 20$.
However, the behaviour of the relative error with $m$ is non-monotonic and small errors are also obtained for the cases when $m=14$.
Since the computational cost of the DMD accelerations goes with $m^2$, the number of operations necessary to performing the DMD iterations with $m=14$ is $0.49$ times of the one needed to perform the iterations with $m=20$.
Therefore, $m=14$ has been selected, trying to introduce the minimum possible computational overhead by adding the DMD iterations.
Once $m=14$ selected, we look in Figure~\ref{fig:TTsensitivity} to the minimum mean relative error on the $m=14$ row. This one is produced at $s=55$.
Therefore, we select $s=55$ for the DMD process.

\begin{figure}
    \centering
    \includegraphics[scale=0.45]{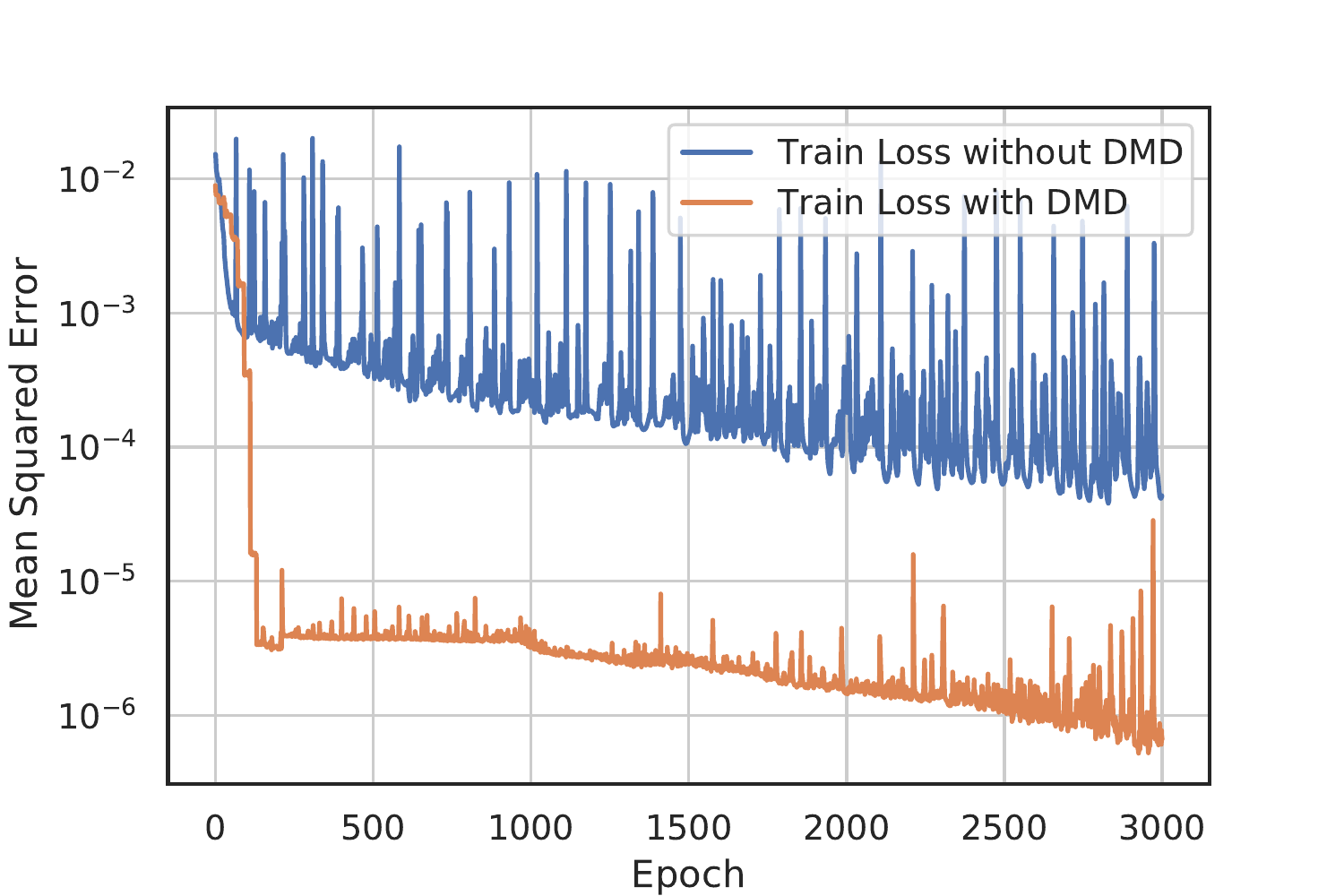}
    \includegraphics[scale=0.45]{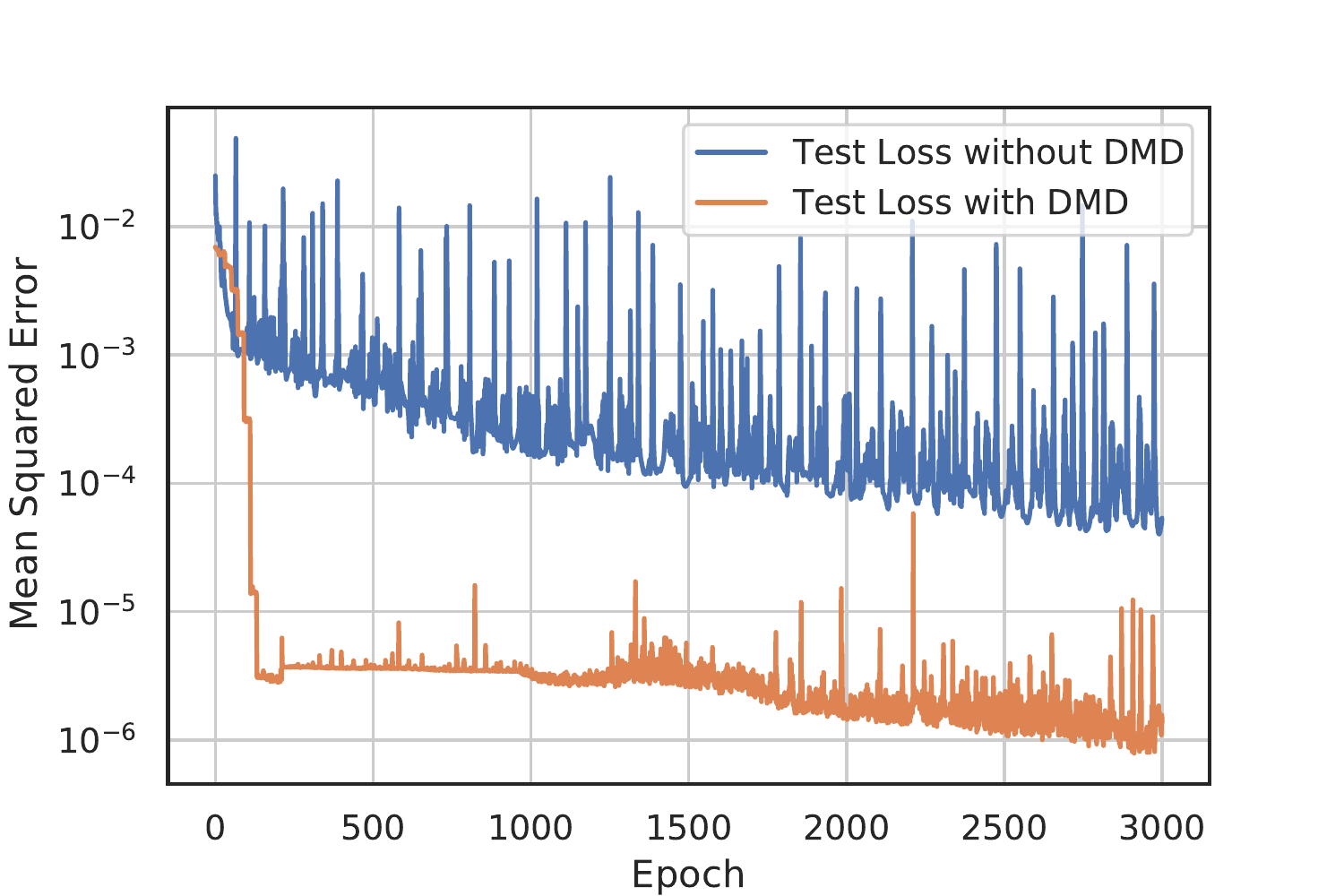}
    \caption{Comparison for the mean squared error with and without DMD acceleration for the training set (left) and the testing set (right).}
    \label{fig:TTperformance}
\end{figure}

The training and validation losses during training using $m=14$ and $s=55$ is presented in Figure~\ref{fig:TTperformance}.
It is immediately clear from this figure the superior performance of the DNN trained with DMD iterations.
In fact, the DNN trained with DMD iterations outperforms the ones trained without these ones by approximately two decades in both the training and validation sets.

By analyzing the training and test losses in Figure~\ref{fig:TTperformance},the effect of DMD iterations is seen very clearly for the first steps.
Large reductions in mean squared error are observed when the DMD iterations are triggered.
Note that, implicitly, the learning rate of DMD iterations is $1.0$.
This explains the large reduction observed in the error.
Then, as more epochs are performed, the reduction of the error due to the DMD iterations becomes less marked.
This is because the large steps taken in the DMD optimization process are less performant when mean squared errors are already small.
The reduction in the performance of DMD for small errors suggest that annealing or relaxation are necessary when performing the DMD iterations in Algorithm~\ref{algo:DMDpBP}.
We are currently targeting this problem.

Further analysis of Figure~\ref{fig:TTperformance} shows that a similar improvement in the performance than the one obtained for the training set is also obtained for the testing set.
This confirms the hypothesis noted in the sensitivity study, which suggested that DMD iterations do not introduce overfitting effects during the optimization process.
Nonetheless, this may be an artifact of the regression problems selected.
We suggest limiting the number of DMD steps $s$ if overfitting issues are found when applying this technique to other regression problems.

When analyzing the mean squared errors in Figure~\ref{fig:TTperformance}, it is observed that the stochastic oscillations in the loss functions are damped when using the DMD iterations.
This is because only the principal modes of the weights are used when computing the DMD evolution.
Thus, stochastic discrepancies between weights at each layer that results in a stochastic behaviour of the loss function are filtered out by DMD.
Once again, this is beneficial for the present regression problem but it is not clear that it will be beneficial for every single regression problem.
If by flattening out the stochastic behaviour of weights the performance of the DNN is deteriorated, we suggest to include add a random noise at the end of the DMD iterations.
To keep consistency, these noise should be produced by randomly sampling the difference between the distributions of weights obtained after the DMD process and the original one obtained at step $m$.

As a final comment, the wall time necessary to train the DNN with DMD iterations was $1.41$ times larger than the one necessary to train the DNN without DMD iterations.
This is larger of the $1.07$ factor that can be theoretically predicted by analyzing the number of operations.
Besides the extra cost of computing the DMD iterations, an important part of the extra computational cost was produced when extracting the weights from the DNN after every backpropagation iteration and when assigning the weights back to the DNN after the DMD iterations.
Native implementation of this method in the machine learning library (which was TensorFlow~\cite{tf2016} in this case) can greatly alleviate this extra computational cost.
Additionally, the DMD iterations are currently not implemented for GPU architectures.
Their implementation in these architectures can bring the wall-time difference by adding DMD iterations closer to the theoretical limit.
Besides, if acceleration is required with the current implementation of the code, stopping the training of the DNN after a few hundred of iterations can provide important acceleration, while still maintaining a good training performance in the DNN.
%
%%%%%%%%%%%%%%%%%%%%%%%%%%%%%%%%%%%%%%%%%%%%%%%%%%%%%%%%%%%%%%%%%%%%%%%%%%%%%%
\section{Conclusion and Perspectives}
%%%%%%%%%%%%%%%%%%%%%%%%%%%%%%%%%%%%%%%%%%%%%%%%%%%%%%%%%%%%%%%%%%%%%%%%%%%%%%
In this paper we have introduced a method form accelerating the training of Deep Neural Networks.
This method is based on the observation that, for regression problems, the distributions of weights per layer follow approximately a monotonic path to convergence.
Thus, Dynamic Mode Decomposition (DMD) is used to infer these dynamics and to compute the evolution of weights per layer.
The computational cost of the DMD method proposed is significantly less than the one of backpropagation.
Additionally, iterations are not necessary when computing the DMD evolution and the estimated converged state for weights can be computed with one step.
This method has been applied for optimizing a DNN used as a surrogate for the transport of a reactive pollutant in the atmosphere.
Owe to a large sensitivity to input parameters and a wide range of solutions that must be represented with the DNN in this regression problem, backpropogation performs rather poorly when training this network.
In particular, a slow convergence rate for weights is observed with a large extent of noise during training.
It is observed that mixing the DMD methods with backpropagation not only reduces the mean squared error during training by a factor of $~100$, but also largely mitigate the noise observed when trianing the ANN.

In the future, we will be applying this improved training method to other regression problems of interest for the scientific machine learning community.
We anticipate this methodology to assistant in rapid prototyping of DNN architecture and hyperparameter optimization studies for regression problems by reducing training times while improving training accuracy.
Additionally, non-linear manifold representations will be replacing the linear Singular Value Decomposition process in DMD in an attempt to better capture the nonlinear effects in the evolution of weights that lead to larger errors when increasing the number of DMD steps.
This will ultimately result in a better performance for DMD acceleration, which yields a faster training process.
Finally, beyond our applications, we think that it would be interesting for the regression community to try this method on different problems.
Our hope is that by repeatedly testing this method, bottlenecks will be identified, allowing us to introduce improvements in this methodology.

%\newpage
\section*{Broader Impact}

In the current work we propose a method to accelerate the training of Deep Neural Networks (DNN).
This method can lead to a significantly faster training DNNs, which makes machine learning accessible to groups without high performance computing capabilities.
Additionally, it opens the possibility of training larger DNNs to groups that do have these capabilities.
The particular application of each trained DNN exceeds, evidently, the scope of this work.

Nonetheless, in this work we demonstrated that DNNs can be trained to rapidly predict the spread of a pollutant in an emergency response scenario.
This application is potentially very valuable to orchestrate a prevention/palliative response that can minimize the impact on the surrounding population.
Additionally, it opens the possibility of the more intensive usage of DNNs to better understand earth science and climate.
We believe that the application of DNNs for less noble causes can be limited by strictly focusing our applications in purely scientific examples.

%removed for submission
\section*{Acknowledgements}

The research activities of the authors at Texas A\&M has been made possible through a grant by the Department of the Defense, Defense Threat Reduction Agency under Award No. HDTRA1-18-1-0020. 

The research activities of G.D.P. are supported by a Los Alamos National Laboratory LDRD project entitled "Machine Learning for Turbulence (MELT)" \#20190059DR and by the National Nuclear Security Administration Office of Experimental Sciences (NNSA OES) via a project entitled ``Imaging for Radiography''.

The content of the information does not necessarily reflect the position or the policy of the federal government, and no official endorsement should be inferred. This document has been approved by Los Alamos National Laboratory for public release as LA-UR-20-24004.

\bibliographystyle{plain} % We choose the "plain" reference style
\bibliography{biblio}

\section*{Appendix 1: Mathematical description of dispersion of reactive pollutants}

\begin{figure}[H]
    \centering
    \includegraphics[scale=0.7]{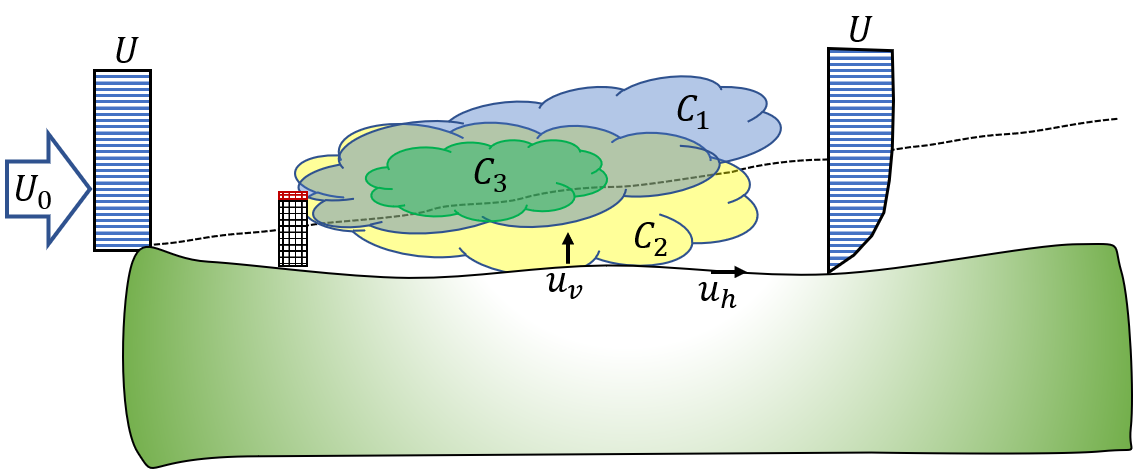}
    \caption{Schematic description of the problem}
    \label{fig:problem_description}
\end{figure}

A schematic description of the reactive pollutant problem is presented in Figure~\ref{fig:problem_description}.
In this setting, boundary layer flow profiles develops over an irregular terrain.
Non-dimensionalized quantities are assumed throughout the formulation of the problem.
The incident wind velocity to the terrain is $U_0$ and, because of the irregularities, horizontal $u_h$ and vertical $u_v$ slip velocities are produced next to this one.
Note that in reality the velocity at ground level must be zero.
However, these slip velocities are modeling the velocities tenths of centimeters away from the ground.

At some point in the terrain there is a chimney emanating to solute reactants $1$ and $2$.
The concentration of these two reactants through the domain are $c_1$ and $c_2$ respectively.
Both this reactants diffuse from the emission point, with a coefficient $D$, generating a cloud and are advected by the background velocity profile.
Additionally, reactants $1$ and $2$ react to produce solute $3$ with a first order reaction constant $K_{12}$.
The concentration field of $3$ throughout the domain is names $c_3$.

Assuming that the solutes have no effect in the external flow field allows to separate the modeling of the flow field from that of the solutes.
The development of the flow field over the terrain is modeled by a boundary layer profile.
Within the boundary layer approximations, we can propose a self-similar solution for the 2D velocity field $\vec{u} = (u_x, u_y)$ as follows:
\begin{equation}
\begin{split}
        u_x(x,y) = f'(\eta) U_0 \,, \\
        u_y(x,y) = \frac{1}{2} \frac{\nu U_0}{x} \left[ \eta f'(\eta) - f(\eta) \right] \,,
\end{split}
\end{equation}
where $\eta = \sqrt{U_0/(2 \nu x)} y$ is the self-similar variable, $\nu = 10^{-5}$ is the kinematic viscosity of air, and $f(\eta)$ is the characteristic solution to the flow profile that is solved by a Blasius equation.
Assuming horizontal slip next to the ground, $u_x(x,y=0) = u_h$, and a vertical inflow profile of the form $u_y(x, y=0) = u_v/\sqrt{x}$, the Blasius problem is posed as follows:
\begin{equation}
    \left\{
    \begin{array}{c}
         2 f''' + f'' f = 0 \,, \\
         f'(\eta = 0) = \frac{u_h}{U_0} \,,\\
         f(\eta = 0) = -\frac{2 u_v}{\sqrt{\nu U_0}} \,,\\
         f'(\eta \rightarrow \infty) = 1 \,.
    \end{array}
    \right.
\end{equation}

\begin{figure}
    \centering
    \includegraphics[scale=0.5]{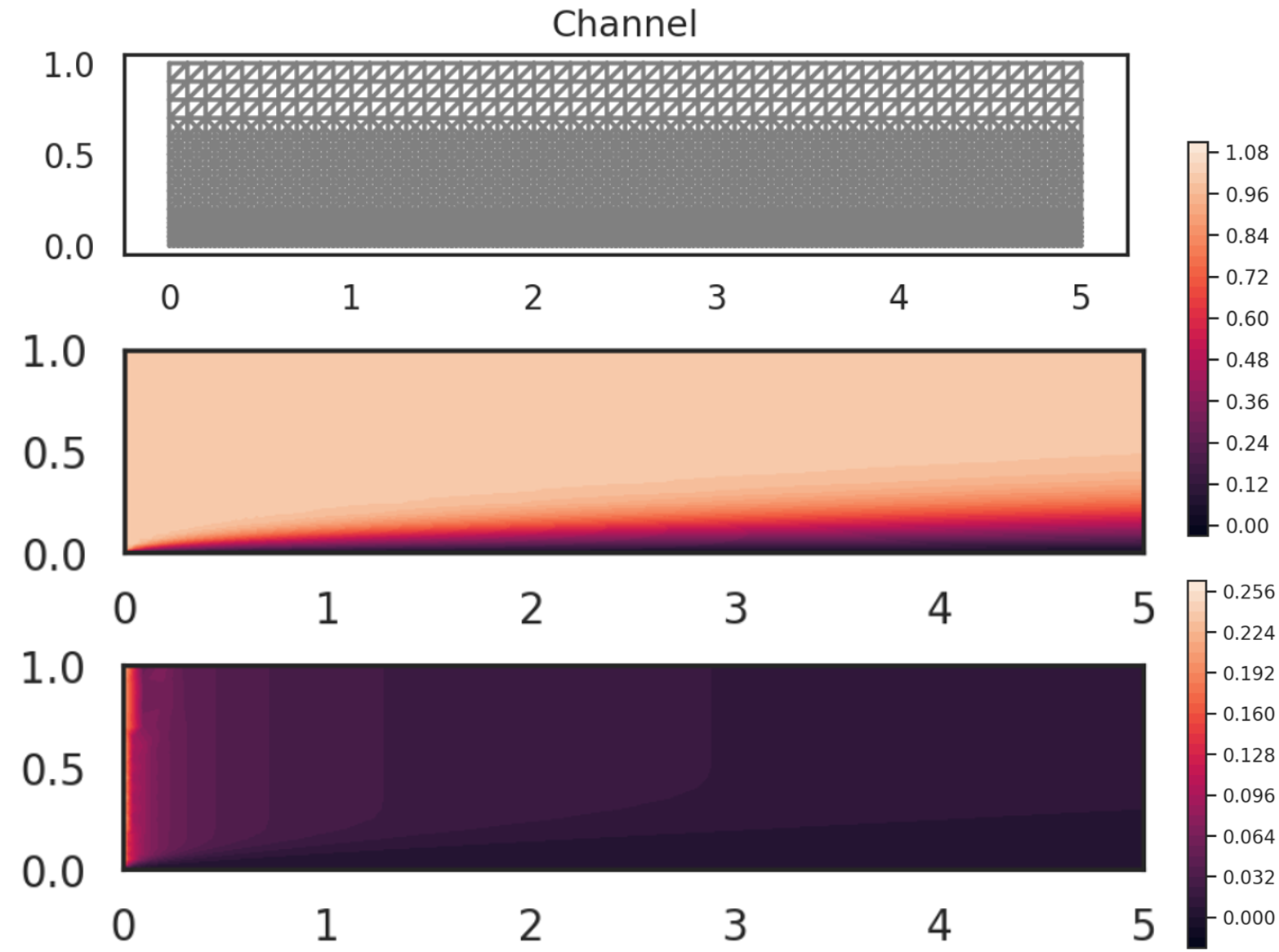}
    \caption{Example of the background velocity profile. {\it Top}: finite element mesh. {\it Center}: horizontal velocity profile $u_x$. {\it Bottom}: vertical velocity profile $u_y$.}
    \label{fig:BL_profile}
\end{figure}

This system is solved numerically using the shooting method.
The resulting velocity profile is then applied to a finite element mesh, which is shown along with the horizontal and vertical velocity profile in Figure~\ref{fig:BL_profile}.
It is observed in this figure how the velocity profile reduces its speed next to the ground, which leads to a vertical velocity profile away from the terrain.

Once the background velocity fixed, the next step consists of modeling the transport and reaction of solutes $1$, $2$ and $3$.
Hence, the concentration field $c_1$ for these solutes is modeled by the following system of coupled partial differential equations:
\begin{equation}
    \left\{ 
    \begin{array}{c}
         \vec{u} \cdot \nabla c_1 - D \nabla^2 c_1 - K_{12} c_1 c_2 - Q_1 = 0 \,,  \\
         \vec{u} \cdot \nabla c_2 - D \nabla^2 c_2 - K_{12} c_1 c_2 - Q_2 = 0 \,,  \\
         \vec{u} \cdot \nabla c_3 - D \nabla^2 c_3 + K_{12} c_1 c_2 + K_3 c_3= 0 \,,
    \end{array}
    \right.
\end{equation}
subjected to Neumann boundary conditions at the terrain and inflow-outflow boundary conditions elsewhere.
$Q_1$ and $Q_2$ are the source of solutes $1$ and $2$ respectively, which represents the emissions from the chimney.
These ones are modeled as follows:
\begin{equation}
    \begin{split}
    Q_1 =
    \left\{ 
    \begin{array}{ccc}
         0.1 & \text{if} & (x-0.1)^2+(y-0.1)^2 < 0.25 ,\\
         0.0 & \text{otherwise} ,
    \end{array}
    \right. \\
    Q_2 =
    \left\{ 
    \begin{array}{ccc}
         0.1 & \text{if} & (x-0.1)^2+(y-0.3)^2 < 0.25 ,\\
         0.0 & \text{otherwise} .
    \end{array}
    \right.
    \end{split}
\end{equation}

\begin{figure}
    \centering
    \includegraphics[scale=0.5]{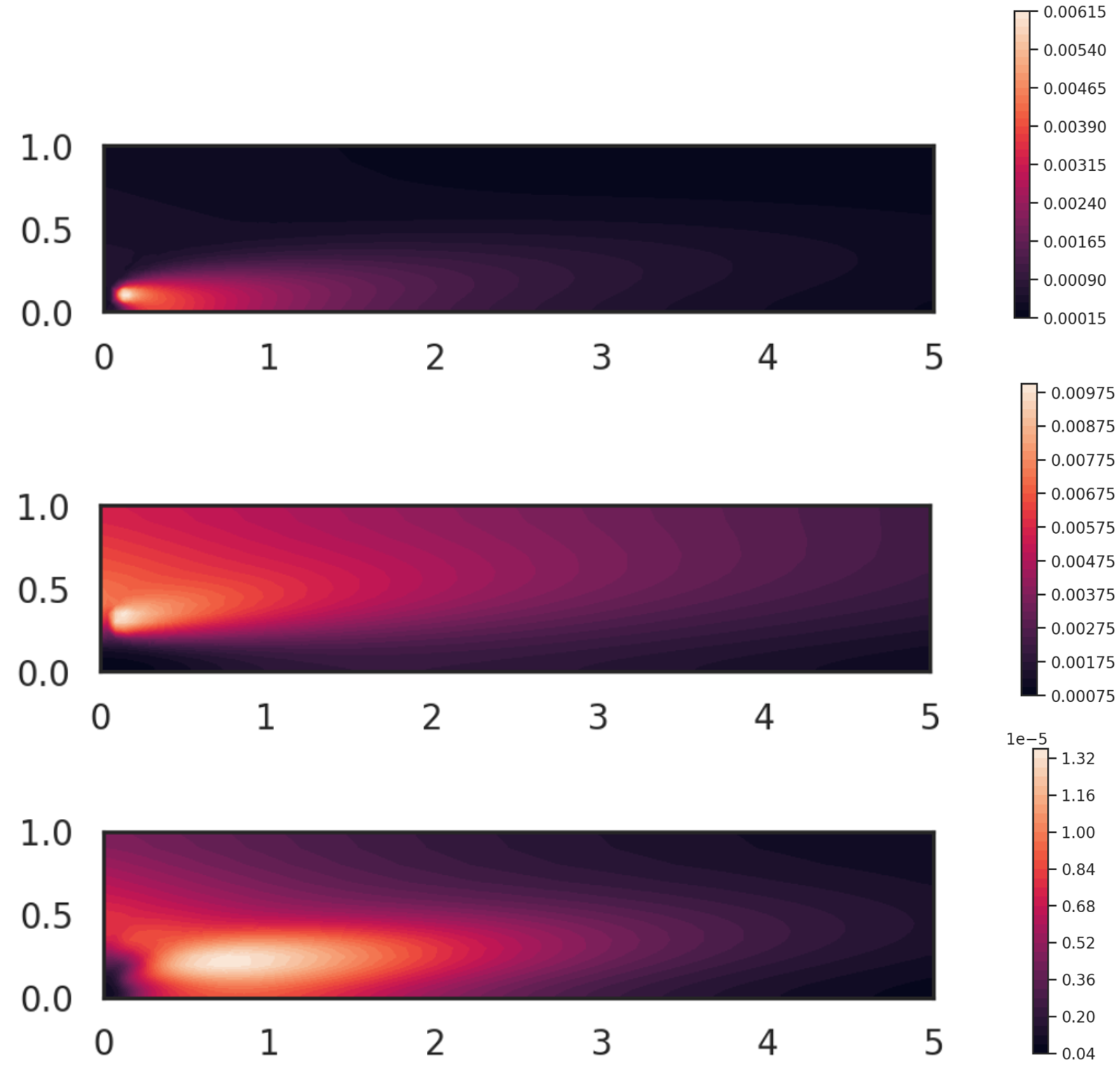}
    \caption{Examples of concentration profiles obtained for solutes $1$ ({\it top}), $2$ ({\it center}), and $3$ ({\it bottom}).}
    \label{fig:concentration_profiles}
\end{figure}

The solution of the resulting system is performed using mixed finite elements with common bases for all solutes.
Examples of the solutions obtained for the three solutes' concentration fields are presented in Figure~\ref{fig:concentration_profiles}.

\end{document}